\documentclass[conference]{IEEEtran}
\IEEEoverridecommandlockouts
\usepackage{cite}
\usepackage{amsmath,amssymb,amsfonts}
\usepackage{algorithmic}
\usepackage{graphicx}
\usepackage{textcomp}
\usepackage{xcolor}
\def\BibTeX{{\rm B\kern-.05em{\sc i\kern-.025em b}\kern-.08em
    T\kern-.1667em\lower.7ex\hbox{E}\kern-.125emX}}
\usepackage{hyperref}
\usepackage{mathtools}
\usepackage{layouts}
\usepackage{graphicx}
\usepackage{multirow}
\usepackage{amsmath}
\usepackage{amssymb}
\usepackage{array}
\usepackage{tabu}
\usepackage{array}
\usepackage{stackengine}
\usepackage{booktabs}

\newcommand{\tran}{^\intercal}
\newcommand{\inv}{^{\text{-}1}}
\usepackage[]{algorithm2e}

\usepackage{amsmath,amsfonts,bm}

\def\eqref#1{equation~\ref{#1}}
\def\Eqref#1{Equation~\ref{#1}}

\def\1{\bm{1}}

\def\vtheta{{\bm{\theta}}}
\def\va{{\bm{a}}}

\def\vf{{\bm{f}}}
\def\vg{{\bm{g}}}

\def\vl{{\bm{l}}}
\def\vm{{\bm{m}}}

\def\vp{{\bm{p}}}
\def\vq{{\bm{q}}}

\def\vs{{\bm{s}}}

\def\vu{{\bm{u}}}
\def\vv{{\bm{v}}}

\def\vx{{\bm{x}}}

\def\mI{{\bm{I}}}
\def\mJ{{\bm{J}}}
\def\mK{{\bm{K}}}

\def\mM{{\bm{M}}}

\def\mR{{\bm{R}}}

\def\mT{{\bm{T}}}

\DeclareMathAlphabet{\mathsfit}{\encodingdefault}{\sfdefault}{m}{sl}
\SetMathAlphabet{\mathsfit}{bold}{\encodingdefault}{\sfdefault}{bx}{n}

\DeclareMathOperator*{\argmin}{arg\,min}

\begin{document}

\title{Differentiable Physics Models for Real-world\\Offline Model-based Reinforcement Learning}

\author{\IEEEauthorblockN{Michael Lutter$^{*}$, Johannes Silberbauer$^{*}$\thanks{$^{*}$ Equal Contribution} , Joe Watson, Jan Peters\thanks{This project has received funding from the European Union’s Horizon 2020 research and innovation program under grant agreement No \#640554 (SKILLS4ROBOTS). Furthermore, this research was also supported by grants from ABB and NVIDIA.}
}
\IEEEauthorblockA{\textit{Computer Science Department}, 
\textit{Technical University of Darmstadt}\\
\{michael, joe, jan\}@robot-learning.de}
}

\refstepcounter{figure}


\makeatletter
\let\@oldmaketitle\@maketitle
\renewcommand{\@maketitle}{\@oldmaketitle
  \includegraphics[width=\linewidth]{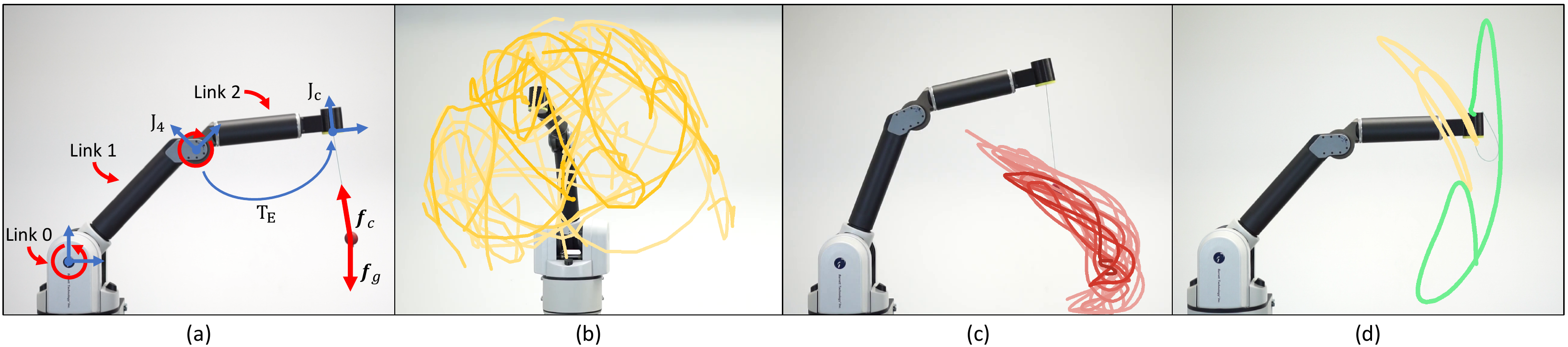} \\[0.15em]
  \footnotesize{Fig.~\thefigure. (a) The identified dynamics (red) and kinematic (blue) parameter of the Barrett WAM for the Ball in a Cup task. (b) Exploration data for the DiffNEA to infer the robot dynamics parameters. (c) Exploration data for the DiffNEA white-box model to infer $T_{E}$ and the string length. (d) Successful swing-up on real system using offline model based reinforcement learning.}
  \label{fig:title} \medskip \vspace{-10pt}}
\makeatother

\maketitle

\thispagestyle{empty}
\pagestyle{empty}


%

\begin{abstract}
A limitation of model-based reinforcement learning (MBRL) is the exploitation of errors in the learned models. 
Black-box models can fit complex dynamics with high fidelity, but their behavior is undefined outside of the data distribution.
Physics-based models are better at extrapolating, due to the general validity of their informed structure, but underfit in the real world due to
the presence of unmodeled phenomena.
In this work, we demonstrate experimentally that for the \emph{offline} model-based reinforcement learning setting, physics-based models
can be beneficial compared to high-capacity function approximators if the mechanical structure is known.
Physics-based models can learn to perform the ball in a cup (BiC) task on a physical manipulator using only 4 minutes of sampled data using offline MBRL.
We find that black-box models consistently produce unviable policies for BiC as all predicted trajectories diverge to physically impossible state, despite having access to more data than the physics-based model. In addition, we generalize the approach of physics parameter identification from modeling holonomic multi-body systems to systems with nonholonomic dynamics using end-to-end automatic differentiation.\\
Videos: 
\textbf{\href{https://sites.google.com/view/ball-in-a-cup-in-4-minutes/}{https://sites.google.com/view/ball-in-a-cup-in-4-minutes/}}
\end{abstract}


\section{Introduction} 
\noindent The recent advent of model-based reinforcement learning has sparked renewed interest in model learning \cite{lutter2018deep, Lutter2019Energy, greydanus2019hamiltonian, gupta2019general, cranmer2020lagrangian, saemundsson20variational, zhong2019symplectic}.
A learned model should reduce the sample complexity of the reinforcement learning task, through interpolation and extrapolation of the acquired data, and thus enable the application to physical systems.
Building upon the vast literature of model learning for control, various new approaches to improve black-box models with physics have been proposed. 
However, the question of what is a good model for MBRL and how this might differ from models for control has not been thoroughly addressed.
A popular opinion is that black-box models are preferable, as such models are applicable to arbitrary systems and can approximate complex dynamics with high fidelity.
In contrast, physics-based models can underfit due to unmodeled phenomena and 
require specific domain knowledge about the system.   

\medskip
\noindent In this work, we discuss the challenges of model learning for MBRL and contrast them to the challenges of model-based control synthesis, the original motivation for model learning.
We compare these requirements to the characteristics of black-box and physics-based models.
To experimentally highlight the differences between model representations for MBRL, we compare the performance of each model type using offline MBRL applied to the common RL benchmark of ball in a cup (BiC) \cite{kober2009policy, schwab2019simultaneously, klink2019self} on the physical Barrett WAM.
The model performance is evaluated using offline MBRL as this approach is the most susceptible for model exploitation and hence amplifies the differences between model representations.
BiC on the Barrett WAM is a challenging task for MBRL as the task requires precise movements, combines various physics phenomena including cable drives, rigid-body-dynamics and string dynamics and uses reduced and maximal coordinates.

\medskip
\noindent In the process we extend the identification of physics models to nonholonomic systems \cite{bloch2003nonholonomic}, which previously were limited to multi-body kinematic chains \cite{atkeson1986estimation, diffNEA, sutanto2020encoding}. Using the advancements in automatic differentiation (AD) \cite{Rall81} and careful reparametrizations of the physics parameters one can infer a guaranteed physically plausible model for arbitrary mechanical systems with unconstrained gradient based optimization - if the kinematic structure is known. Thus this extension generalizes the elegantly crafted features of \cite{atkeson1986estimation} by backpropagating through the computational graph spanned by the differential equations of physics.  

\medskip
\noindent \textbf{Contributions} We provide a experimental evaluation of different model representations for solving BiC with offline MBRL. We show that for some tasks, e.g., BiC, guaranteed physically plausible models are preferable compared to deep networks despite the inherent underfitting. Physics-based white-box models, learned with only four minutes of data, are capable of solving BiC with offline MBRL. Deep network models do not achieve this task. In addition, we extend the existing methods for physics parameters identification to systems with maximal coordinates and nonholonomic inequality constraints.

\medskip
\noindent In the following we discuss the challenges of models for MBRL (Section \ref{sec:models_for_mbrl}), describe our approach to physically plausible parameter identification for systems with holonomic and nonholonomic constraints (Section \ref{sec:model}). Finally, Section \ref{sec:experiments}
experimentally compares the model representations extensively by applying them to offline MBRL to solve the BiC task on the real Barrett WAM with three different string lengths .

\section{Model Representations} \label{sec:models_for_mbrl}
\noindent Model learning, or system identification \cite{aastrom1971system}, aims to infer the parameters $\vtheta$ of the system dynamics from data containing the system state $\vx$ and the control signal $\vu$.
In the continuous time case the dynamics are described by
\begin{gather}
    \ddot{\vx} = f(\vx, \dot{\vx}, \vu; \vtheta).
\end{gather}
The optimal parameters $\vtheta^{*}$ are commonly obtained by minimizing the error of the forward or inverse dynamics model,
\begin{align} \label{eq:loss}
    \vtheta^{*}_{\text{for}} &= \argmin_{\vtheta}
    \textstyle\sum_{i=0}^{N} \| \ddot{\vx}_i - \hat{\vf} \left(\vx_i, \dot{\vx}_i, \vu_i; \vtheta \right)\|^2 \\
    \vtheta^{*}_{\text{inv}} &= \argmin_{\vtheta}
    \textstyle\sum_{i=0}^{N} \| \vu_i - \hat{\vf}^{\text{-}1}\left(\vx_i, \dot{\vx}_i, \ddot{\vx}_i; \vtheta \right)\|^2.
\end{align}
Depending on the chosen representation for $\vf$, the model hypotheses spaces and the optimization method changes.

\medskip
\noindent \textbf{White-box Models} These models use the analytical equations of motions to formalize the hypotheses space of $\vf$ and the interpretable physical parameters such as mass, inertia or length as parameters $\vtheta$. Therefore, white-box models are limited to describe the phenomena incorporated within the equations of motions but generalize to unseen state regions as the parameters are global. 
This approach was initially proposed for rigid-body chain manipulators by Atkeson et. al. \cite{atkeson1986estimation}. Using the recursive Newton-Euler algorithm (RNEA) \cite{Featherstone2007rigid}, the authors derived features that simplify the inference of $\vtheta$ to linear regression. 
The resulting parameters must not be necessarily be physically plausible as constraints between the parameters exist. For example, the inertia matrix contained in $\vtheta^{*}$ must be positive definite matrix and fulfill the triangle inequality. Since then, various parameterizations for the physical parameters have been proposed to enforce these constraints through the virtual parameters.
Various reparameterizations \cite{traversaro2016identification, wensing2017linear, sutanto2020encoding} were proposed to guarantee physically plausible inertia matrices.
Using these virtual parameters, the optimization does not simplify to linear regression but can be solved by unconstrained gradient-based optimization and is guaranteed to preserve physically plausibility.  

\medskip
\noindent \textbf{Black-box Models} These models are generic function approximators such as locally linear models \cite{Atkeson_AIR_1998, schaal2002scalable}, Gaussian processes  \cite{kocijan2004gaussian, nguyen2009model, nguyen2010using}, deep- \cite{jansen1994learning, sanchez2018graph} or graph networks \cite{sanchez2020learning} for $\vf$.
These approximators can fit arbitrary and complex dynamics with high fidelity but have an undefined behavior outside the training distribution and might be physically unplausible even on the training domain.
Due to the local nature of the representation, the behavior is only well defined on the training domain and hence the learned models do not extrapolate well.
Furthermore, these models can learn implausible system violating fundamental physics laws such as energy conservation. Only recently deep networks were augmented with knowledge from physics to constrain network representations to be physically plausible on the training domain \cite{lutter2018deep, Lutter2019Energy, greydanus2019hamiltonian, gupta2019general, cranmer2020lagrangian, saemundsson20variational, zhong2019symplectic}.
However, the behavior outside the training domains remains unknown. 

\section{Models for Model-Based RL}
\noindent
For MBRL, black-box models have been widely adopted due to their generic applicability and simplicity \cite{janner2019trust, chua2018deep, langlois2019benchmarking}.
In the following, we will elaborate on specific aspects of MBRL which make model learning for MBRL challenging,
and questions the use of black-box over white-box structures.

\medskip
\noindent 
\textbf{Data Distribution} MBRL is commonly applied to complex tasks  which involve contacts of multiple bodies, such as object manipulation and locomotion.
In this case, the training data lies on a complex manifold separating physically feasible and impossible states, e.g., object contact vs. penetration.
In addition, the data is not uniformly distributed over the set of feasible states, but accumulated at the manifold boundaries. 
In the considered BiC task, the ball is mostly observed at a certain distance from the cup due to the string constraint, rarely closer and never further.
This complex data manifold is in contrast to model learning for
simper tasks 
where the data is evenly distributed in the feasible state region, which is the convex set of the training data. 

\medskip
\noindent
\textbf{Model Usage} MBRL uses the model to plan trajectories and evaluate the policy.
During the planning, the predicted trajectories can venture to physically impossible states and exploit potential shortcuts to improve control.
This behavior is especially likely in 
constrained
tasks where one needs to traverse along the edge of the feasible states.
For example, to solve the BiC task, one needs to plan with the string maximally extended.
In this configuration, the planned trajectory can easily diverge to states where the string-length would be longer than physically possible.
Conversely, for model-based policies such impossible regions are no concern for the model.
In this setting the model is not queried in these configurations as the system cannot enter these states without system damage, malfunction or erroneous measurements. 

\medskip
\noindent
These two characteristics of MBRL affect the model representations differently.
Black-box models are less adapt at learning models from highly localized data as they can only extrapolate locally.
In particular, this local interpolation can fail at the boundaries where bodies are in contact.
Ill-fitted boundaries make it very likely that the planned trajectories diverge to physically implausible regions and that the policy optimization exploits any shortcuts within these regions.
More data cannot resolve the problem, as the data from the physically implausible regions cannot be obtained from the real-world system.
In contrast, white-box models are less susceptible to the irregular data distribution due to the global structure.
Furthermore, many strategies have been developed for white-box models to avoid physically implausible regions within the simulation community.
For example, white-box models avoid implausible states by generating forces orthogonal to the violated constraint to push the system state back to the physically feasible states.
Due to these advantages of white-box models, this model representation can be beneficial for MBRL applications and the underfitting, which limits the application to model-based policies, is only of secondary concern.
To test this hypothesis, we consider the BiC task, which relies heavily on the string constraint. In the following we construct a generic differentiable white-box structure for such a nonholonomic constraint expressed in maximal coordinates.

\section{Differentiable Simulation Models}
\label{sec:model} 
\noindent In the following two sections we describe the used differentiable simulator based on the Newton-Euler algorithm in terms of the elegant Lie algebra formulation \cite{kim2012lie}.
First we describe the simulator for systems with holonomic constraints, i.e., kinematic chains, and then extend it to systems with nonholonomic constraints. In the following we will refer to these models as \emph{DiffNEA} as these models are based on the differentiability of the Newton-Euler equation.

\medskip
\noindent \textbf{Rigid-Body Physics with Holonomic Constraints}
For rigid-body systems with holonomic constraints the system dynamics can expressed analytically in maximal coordinates $\vx$, i.e.,  task space, and reduced coordinates $\vq$, i.e., joint space.
If expressed using maximal coordinates, the dynamics is a constrained problem with the holonomic constraints $g(\cdot)$.
For the reduced coordinates, the dynamics are reparametrized such that the constraints are always fulfilled and the dynamics are unconstrained.
Mathematically this is described by
\begin{gather}
    \ddot{\vx} = f(\vx, \dot{\vx}, \vu; \vtheta) \hspace{15pt}
    \text{s.t.} \hspace{15pt} 
    g(\vx; \vtheta) = 0 \\
    \Rightarrow
    \ddot{\vq} = f(\vq, \dot{\vq}, \vu; \vtheta).
\end{gather}
For model learning of such systems one commonly exploits the reduced coordinate formulation and minimizes the squared loss of the forward or inverse model.
For kinematic trees the forward dynamics $\vf(\cdot)$ can be easily computed using the articulated body algorithm (ABA) and the inverse dynamics $\vf\inv(\cdot)$ via the recursive Newton-Euler algorithm (RNEA)~\cite{Featherstone2007rigid}. Both algorithms are inherently differentiable and one can solve the optimization problem of \Eqref{eq:loss} using backpropagation.

\medskip
\noindent In this implementation, we use the Lie formulations of ABA and RNEA \cite{kim2012lie} for compact and intuitive compared to the initial derivations by \cite{atkeson1986estimation, Featherstone2007rigid}.
ABA and RNEA propagate velocities and accelerations from the kinematic root to the leaves and the forces and impulses from the leaves to the root. This propagation along the chain can be easily expressed in Lie algebra by
\begin{align}
    \bar{\vv}_{j} &= \text{Ad}_{\mT_{j,i}}\bar{\vv}_i, &
    \bar{\va}_{j} &= \text{Ad}_{\mT_{j,i}}\bar{\va}_i, \\
    \bar{\vl}_{j} &= \text{Ad}^T_{\mT_{j,i}}\bar{\vl}_i, &
    \bar{\vf}_{j} &= \text{Ad}^T_{\mT_{j,i}}\bar{\vf}_i.
\end{align}
with the generalized velocities $\bar{\vv}$, accelerations $\bar{\va}$, forces $\bar{\vf}$, momentum $\bar{\vl}$ and the adjoint transform $\text{Ad}_{\mT_{j,i}}$ from the $i$th to the $j$th link.
The generalized entities noted by $\bar{.}$ combine the linear and rotational components, e.g., $\bar{\vv} = \left[\vv, \bm{\omega} \right]$ with the linear velocity $\vv$ and the rotational velocity $\bm{\omega}$.
The Newton-Euler equation is described by 
\begin{gather*}
    \bar{\vf}_{\text{net}} = 
    \bar{\mM}\bar{\va} - \text{ad}^*_{\bar{\vv}}\bar{\mM}\bar{\vv},
    \\
    \text{ad}^*_{\bar{\vv}} = 
    \begin{bmatrix}
            [\bm{\omega}]   & \mathbf{0} \\
         [\vv] & [\bm{\omega}]
    \end{bmatrix}, 
    \hspace{10pt}
    \bar{\mM} = 
    \begin{bmatrix}
        \mJ       & m[\vp_m] \\
         m[\vp_m]^T & m\mI
    \end{bmatrix}
\end{gather*}
with the inertia matrix $\mJ$, the link mass $m$, the center of mass offset $\vp_m$. Combining this message passing with the Newton Euler equation enables a compact formulation of RNEA and ABA.
The gradient based optimization also enables the reparametrization of the physical parameters with virtual parameters $\vtheta_{\text{v}}$ that guarantee to be physically plausible \cite{traversaro2016identification, wensing2017linear, sutanto2020encoding}. The exact algorithms of ABA and RNEA in Lie algebra as well as the virtual parametrizations can be found in the appendix.

\medskip
\noindent \textbf{Rigid-Body Physics with Nonholonomic Constraints}
For a mechanical system with nonholonomic constraints, the system dynamics cannot be expressed via an unconstrained equations with reduced coordinates. For the system
\begin{align*} \label{eq:nonholonomic}
    \ddot{\vx} = f(\vx, \dot{\vx}, \vu; \vtheta) \hspace{10pt}
    \text{s.t.}\hspace{10pt}
    h(\vx; \vtheta) \leq 0,\hspace{10pt}
    g(\vx, \dot{\vx}; \vtheta) = 0,
\end{align*}
the constraints are nonholonomic as $h(\cdot)$ is an inequality constraint and $g(\cdot)$ depends on the velocity.
Inextensible strings are an example for systems with inequality constraint, while the bicycle is a system with velocity dependent constraints.
For such systems, one cannot optimize the unconstrained problem directly, but must identify parameters that explain the data and adhere to the constraints. 

\medskip
\noindent The dynamics of the constrained rigid body system can be described by the Newton-Euler equation,
\begin{align}
\bar{\vf}_{\text{net}} &= \bar{\vf}_{g} + \bar{\vf}_{c} + \bar{\vf}_{\vu} = \bar{\mM}\bar{\va} - \text{ad}^*_{\bar{\vv}}\bar{\mM}\bar{\vv}, \\
\Rightarrow \bar{\va} &=  \bar{\mM}\inv \left(\bar{\vf}_{g} + \bar{\vf}_{c} + \bar{\vf}_{\vu} +  \text{ad}^*_{\bar{\vv}}\bar{\mM}\bar{\vv} \right),
\end{align}
where the net force $\bar{\vf}_{\text{net}}$ contains the gravitational force $\bar{\vf}_{g}$, the constraint force $\bar{\vf}_{c}$ and the control force $\bar{\vf}_{\vu}$.
If one can differentiate the constraint solver computing the constraint force w.r.t. to the parameters,
one can identify the parameters $\vtheta$ via gradient descent. This optimization problem can be described by
\begin{equation}
\begin{multlined}
\vtheta^{*} = \argmin_{\vtheta}
\textstyle\sum_{i=0}^{N} \| \bar{\va}_i - \bar{\mM}^{\text{-}1}_{\vtheta} \big(\bar{\vf}_{g}(\vtheta) \\ \hspace{25pt}
+ \bar{\vf}_{c}(\bar{\vx}_i, \bar{\vv}_i; \vtheta) + \bar{\vf}_{\vu} +  \text{ad}^*_{\bar{\vv}_i}\bar{\mM}(\vtheta)\bar{\vv}_i \big)\|^2.
\end{multlined}
\end{equation}
For the inequality constraint, one can to reframe it as an easier equality constraint, by passing the function through a ReLU nonlinearity $\sigma(\cdot)$, so $g(\vx; \vtheta) = \sigma(h(\vx; \vtheta)) = 0$.
From a practical perspective, the softplus nonlinearity provides a soft relaxation of the nonlinearity for smoother optimization.
Since this equality constraint should always be enforced, we can utilize our dynamics to ensure this on the derivative level, so
$g(\cdot) = \dot{g}(\cdot) = \ddot{g}(\cdot) = 0$ for the whole trajectory.
With this augmentation, the constraint may now be expressed as $\vg(\vx,\dot{\vx};\vtheta)=\mathbf{0}$.
The complete loss is described can be described by 
\begin{equation}
\begin{multlined}
\vtheta^{*} = \argmin_{\vtheta} \textstyle\sum_{i=0}^{N} \| \bar{\va}_i - \vf(\bar{\vx}_i, \bar{\vv}_i, \bar{\vu}_i; \vtheta)\|^2 \\ \hspace{30pt}
+ \lambda_{g} \| g(\vtheta) \|^2 + \lambda_{\dot{g}}\| \dot{g}(\vtheta) \|^2 +  \lambda_{\ddot{g}} \| \ddot{g}(\vtheta) \|^2
\end{multlined}
\end{equation}
with the scalar penalty parameters $\lambda_{g}$, $\lambda_{\dot{g}}$ and $\lambda_{\ddot{g}}$.


\vspace{-0.5em}
\section{Related Work} 
\noindent Differentiable simulators have been previously proposed for model-based reinforcement learning \cite{degrave2019differentiable, de2018end} and planning \cite{toussaint2018differentiable}. In these works, the authors focus on the differentiability w.r.t. to the previous state and use the differentiable model to backpropagate in time to optimize policies or plans. Instead, we focus on the differentiability w.r.t. to model parameter and deploy the differentiable model for system identification of robotic system described using reduced and maximal coordinates as well as explicit holonomic and nonholonomic constraints. Such systems are the main interest of MBRL as the common task usually cannot be described using solely unconstrained reduced coordinates. 

\medskip
\noindent To obtain differentiable simulators, the main problem is differentiating through the constraint force solver computing $\vf_{c}$. Various approaches have been proposed, e.g., Belbute-Peres et. al. \cite{de2018end} describe a method to differentiate through the common LCP solver of simulators, Geilinger et. al. \cite{geilinger2020add} describe a smoothed frictional contact model and Hu et al. \cite{hu2019difftaichi} describe a continuous collision resolution approach to improve the gradient computation. In this work we follow the approach of \cite{degrave2019differentiable, heiden2019interactive} and use automatic differentiation to differentiate through the closed form solution of $\vf_c$. For our considered task this closed form solution is possible and we do not need to rely on more complex approaches presented in literature. 


\begin{figure}[t]
    \centering
    \includegraphics[width=\columnwidth]{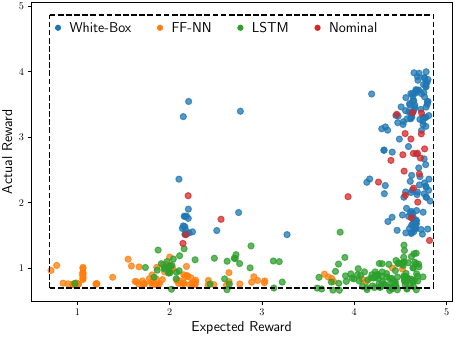}
    \caption{Comparison of the expected reward and the actual reward on the MuJoCo simulator for the LSTM, the feed-forward neural network (FF-NN) as well as the nominal and learnt white-box model. The learnt and nominal white-box model achieve a comparable performance and solve the BiC swing-up for multiple seeds. Neither the LSTM nor the FF-NN achieve a single successful swing-up despite being repeated with 50 different seeds and using all the data of generated by the white-box models.}
    \label{fig:results}
\end{figure} 

\begin{figure*}[t]
    \centering
    \includegraphics[width=\textwidth]{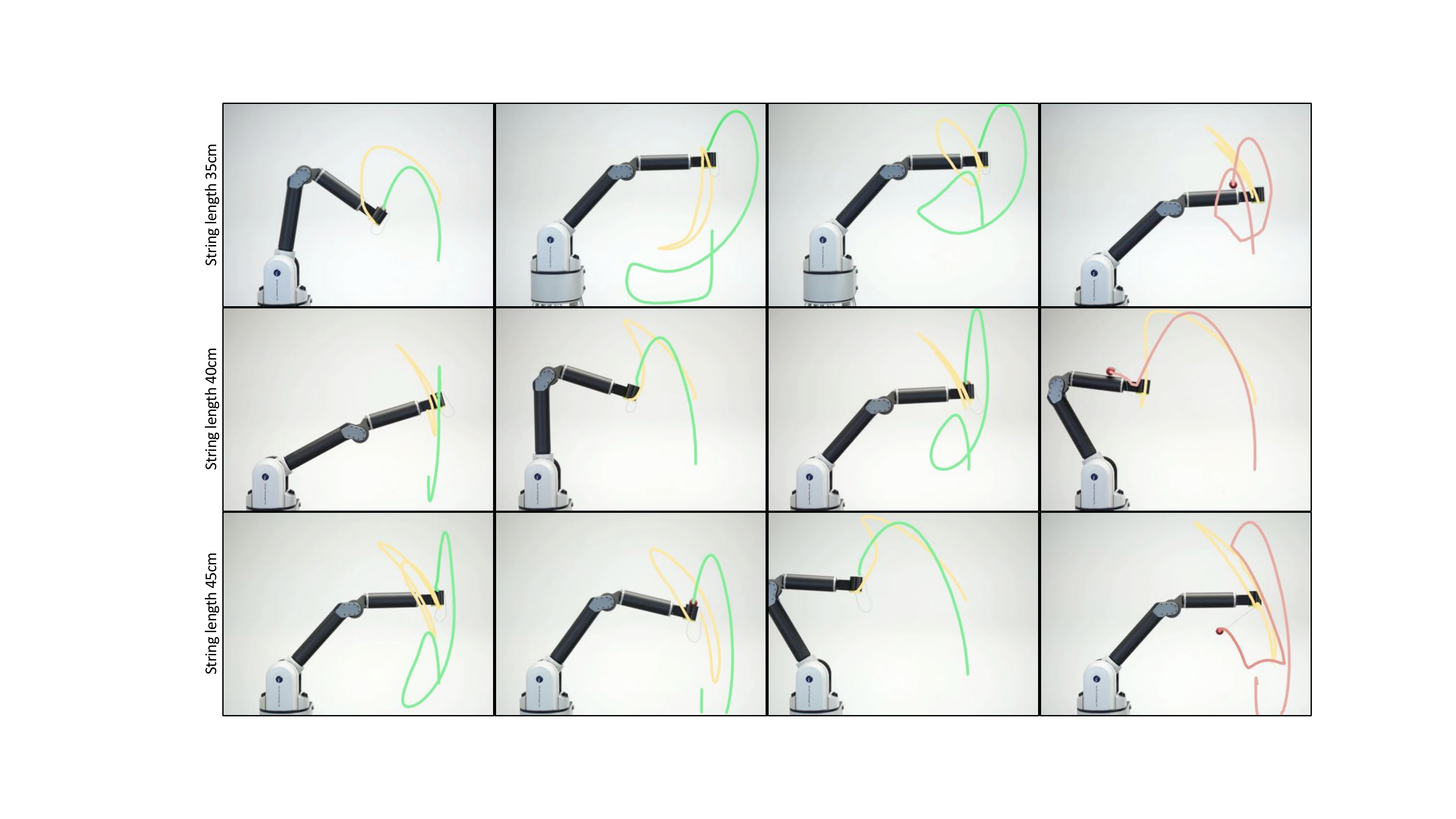}
    \caption{Three different successful swing-ups for the three different string lengths using the DiffNEA White-Box model with eREPS for offline model-based reinforcement learning. This approach can learn different swing-ups from just 4 minutes of data, while all tested black-box models fail at the task. The different solutions are learned using different seeds. The unsuccessful trials of the DiffNEA model nearly solve the BiC tasks but the ball bounces off the cup or arm. 
    Videos and pictures for all models and all experiments can be found at  \textbf{\href{https://sites.google.com/view/ball-in-a-cup-in-4-minutes/}{https://sites.google.com/view/ball-in-a-cup-in-4-minutes/}}}
    \label{fig:movements}
    \vspace{-1.75em}
\end{figure*}

\section{Experimental Setup}
\label{sec:experiments} 
\noindent To evaluate the performance of white-box and black-box models for MBRL, we apply these model representations within an offline MBRL algorithm on the physical system to solve BiC. We test the models within an offline RL algorithm as this approach amplifies the challenges of model learning. In this setting, additional real-world data cannot be used to compensate for modeling errors. BiC is a common benchmark for real-world reinforcement learning and has been used multiple times for model-free reinforcement learning \cite{kober2009policy, schwab2019simultaneously, klink2019self} as well as model-free iterative learning control \cite{bujarbaruah2020learning}.
Until now this task has \emph{not} been solved on a physical system with MBRL as learning a reliable string model is challenging.

\medskip
\noindent 
\textbf{BiC Black-box Model}
\noindent A feedforward network (FF-NN) and a long short-term memory network (LSTM) \cite{hochreiter1997long} is used as black-box model.
The networks model only the string dynamics and receive the task space movement of the last joint and the ball movement as input and predict the ball acceleration, i.e.,
$\ddot{\vx}_B = f(\vx_{J_4}, \dot{\vx}_{J_4}, \ddot{\vx}_{J_4}, \vx_{B}, \dot{\vx}_{B})$. 

\medskip
\noindent 
\textbf{BiC White-box Model}
For this model, the robot manipulator is modeled as a rigid-body chain using reduced coordinates. The ball is modeled via a constrained particle simulation with an inequality constraint.
Both models are interfaced via the task space movement of the robot after the last joint.
The manipulator model predicts the task-space movement after the last joint. The string model transforms this movement to the end-effector frame via $\mT_{E}$ (Figure \ref{fig:title} a), computes the constraint force $\vf_c$ and the ball acceleration $\ddot{\vx}_B$.
Mathematically this model is described by
\begin{gather}
    \ddot{\vx}_B = \textstyle\frac{1}{m_B} \left( \vf_g + \vf_c \right), \\ 
    g(\vx; \vtheta_S) = \sigma(\| \vx_B - \mT_{E} \: \vx_{J_4} \|^2_2 - r^2) = 0,
\end{gather}
where $\vx_B$ is the ball position, $\vx_{J_4}$ the position of the last joint and $r$ the string length. In the following we will abbreviate $\vx_B -\mT_{E} \: \vx_{J_4} = \Delta$ and the cup position by $\mT_{E} \: \vx_{J_4} = \vx_C$. The constraint force can be computed analytically with the principle of virtual work and is described by
\begin{equation}
\begin{gathered}
    \vf_c(\vtheta_S) = - m_B \: \sigma'(z)  \: \frac{\Delta\tran \vg -  \: \Delta\tran \ddot{\mathbf{x}}_C + \: \dot{\Delta}\tran \dot{\Delta}}{\Delta\tran \Delta + \delta} 
\end{gathered}
\end{equation}
with $ z = \| \Delta \|_2 - r,$ and the gravitational vector $\vg$. When simulating the system, we set $\ddot{g} = -\mK_p g - \mK_d \dot{g} \leq 0$ to avoid constraint violations and add friction to the ball for numerical stability. This closed form constraint force is differentiable and hence one does not need to incorporate any special differentiable simulation variants.

\medskip
\noindent 
\textbf{Offline Reinforcement Learning} 
\label{sec:ombrl}
This RL problem formulation studies the problem of learning an optimal policy from a fixed dataset of arbitrary experience \cite{levine2020offline, Lange2012}. Hence, the agent is bound to a dataset and cannot explore the environment. For solving this problem, we use a model-based approach were one first learns a model from the data and then performs episodic model-free reinforcement learning (MFRL) using this approximate model. For the model-free RL we use episodic relative entropy policy search (eREPS) with an additional KL-divergence constraint on the maximum likelihood policy update \cite{ploeger2020high} and parameter exploration \cite{deisenroth2013survey}. The policy is a probabilistic movement primitive (ProMP) \cite{paraschos2013probabilistic, paraschos2018using} describing a distribution over trajectories. 

\medskip
\noindent \textbf{Dataset}
For the manipulator identification the robot executes a $40$s high-acceleration sinusoidal joint trajectory  (Figure \ref{fig:title} b).
For the string model identification, the robot executes a $40$s slow cosine joint trajectories to induce ball oscillation without contact with the manipulator (Figure \ref{fig:title} c).
The ball trajectories are averaged over five trajectories to reduce the variance of the measurement noise.
The training data does not contain swing-up motions and, hence the model must extrapolate to achieve the accurate simulation of the swing-up.
The total dataset used for offline RL contains only $4$ minutes of data. To simplify the task for the deep networks, the training data consists of the original training data plus all data generated by the white-box model during evaluation. Therefore, the network training data contains successful BiC tasks. 

\medskip
\noindent \textbf{Reward}
The dense episodic reward is inspired by the potential of an electric dipole and augmented with regularizing penalties for joint positions and velocities.
The complete reward is defined as
\begin{equation*}
\begin{multlined}
    R(\vs_{<N}) = \exp \left( \frac{1}{2}\max_t \psi_t + \frac{1}{2}\psi_N \right) \\ \hspace{40pt}
    {-} \frac{1}{N} \sum_{i=0}^{N} \lambda_{\vq} \| \vq_i {-} \vq_0 \|^2_2 {+} \lambda_{\dot{\vq}} \| \dot{\vq}_i \|^2_2,
\end{multlined}
\end{equation*}
with $\psi_t = \Delta_t \tran \hat{\vm}(\vq_t) /\left( \Delta_t \tran \Delta_t + \epsilon \right)$ and the normal vector of the end-effector frame
$\hat{\vm}$ which depends on joint configuration $\vq_t$. 
For the white-box model, the \emph{predicted} end-effector frame is used during policy optimization.
Therefore, the policy is optimized using the reward computed in the approximated model. The black-box models uses the true reward, rather than the reward bootstrapped from the learned model.

\begin{table*}[t]
\caption{Offline reinforcement learning results for the ball in a cup task, across both simulation and the physical system. Length refers to the string length in centimeters. Repeatability is reported for the best performing reinforcement learning seed.}
\label{table:results}
  \begin{sc}
  \resizebox{\textwidth}{!}{%
  \begin{tabular}{r cccc cccc}
  	\toprule
  	& \multicolumn{4}{c}{simulation} & \multicolumn{4}{c}{physical system} \\
  	\cmidrule(lr){2-5} \cmidrule(lr){6-9}
	model & length & avg. reward & transferability & repeatability & length & avg. reward & transferability & repeatability \\
    \midrule
    lstm & 40cm & 0.92 $\pm$ 0.37 & 0\% & - & 40cm  & 0.91 $\pm$ 0.56 & 0\% & 0\% \\
    ff-nn & 40cm & 0.86 $\pm$ 0.35 & 0\%  & - & 40cm & 1.46 $\pm$ 0.78 & 0\% & 0\% \\
    Nominal & 40cm & 2.45 $\pm$ 1.15 & \textbf{64\%} & - & 40cm & 1.41 $\pm$ 0.45 & 0\% & 0\% \\
    diffnea & 40cm & \textbf{2.73 $\pm$ 1.64} & 52\%  & -
                             & 40cm & \textbf{1.77 $\pm$ 0.74} & \textbf{60\%} & 90\% \\
                         &&&&& 35cm & 1.58 $\pm$ 0.15 & 30\% & 70\%\\
                         &&&&& 45cm & 1.74 $\pm$ 0.71 & \textbf{60\%} & \textbf{100\%}\\
  \bottomrule
  \end{tabular}
 }
 \end{sc}
\end{table*}


\section{Experimental Results}
\noindent
Videos documenting all experiments can be found at \textbf{\href{https://sites.google.com/view/ball-in-a-cup-in-4-minutes/}{https://sites.google.com/view/ball-in-a-cup-in-4-minutes/}}.

\medskip
\noindent \textbf{Simulation Results}
The simulation experiments test the models with idealized observations from MuJoCo \cite{6386109} and enable a quantitative comparison across many seeds. For each model representation, 15 different learned models are evaluated with 150 seeds for the MFRL. The average statistics of the best ten reinforcement learning seeds are shown in Table \ref{table:results} and the expected versus obtained reward is shown in Figure \ref{fig:results} (a).

\medskip \noindent
The DiffNEA white-box model is able to learn the BiC swing-up for every tested model. The transferability to the MuJoCo simulator depends on the specific seed, as the problem contains many different local solutions and only some solutions are robust to slight model variations. The MuJoCo simulator is different from the DiffNEA model as MuJoCo simulates the string as a chain of multiple small rigid bodies. The performance of the learned DiffNEA is comparable to the performance of the DiffNEA model with the nominal values.   

\medskip \noindent
The FF-NN and LSTM black-box models do not learn a single successful transfers despite being tried on ten different models and 150 different seeds, using additional data that includes swing-ups and observing the real instead of the imagined reward. These learned models cannot stabilize the ball beneath the cup. The ball immediately diverges to a physical unfeasible region. The attached videos compare the real (red) vs. imagined (yellow) ball trajectories. Within the impossible region the policy optimizer exploits the random dynamics where the ball teleports into the cup. Therefore, the policy optimizers converges to random movements. 

\medskip
\noindent \textbf{Real-Robot Results}
The experiments are performed using the Barrett WAM and three different string-lengths, i.e., $35$cm, $40$cm and $45$cm. For each model a 50 different seeds are evaluated on the physical system. A selection of the of trials using the learned DiffNEA white-box model is shown in Figure~\ref{fig:movements}. The roll-outs of the baselines are shown in Figure~\ref{fig:baselines}. The average statistics of the best ten seeds are summarized in Table \ref{table:results}. 

\medskip \noindent 
The DiffNEA white-box model is capable of solving BiC using offline MBRL for all string-lengths. This approach obtains very different solutions that transfer to the physical system. Some solutions contain multiple pre-swings which show the quality of the model for long-planning horizons. The best movements also repeatedly achieve the successful task completion. Solutions that do not transfer to the system, perform feasible movements where the ball bounces of the cup rim. The nominal DiffNEA model with the measured arm and string parameters does not achieve a successful swing-up. The ball always overshoots and bounces of the robot-arm for this model.  

\medskip \noindent 
None of the tested black-box models achieve the BiC swing-up despite using more data and the true rewards during planning. Especially the FF-NN model converges to random policies, which result in ball movement that do not even closely resemble a potential swing-up. The convergence to very different movements shows that the models contain multiple shortcuts capable of teleporting the imagined ball into the cup.

\section{Conclusion \& Future Work} 
\label{sec:conclusion} 
\noindent 
In this paper we argue that for highly constrained tasks, white-box models provide a benefit over black-box model for MBRL, and verify this hypothesis through an extensive evaluation on ball in a cup task on a real robotic platform. The ball in a cup task shows that guaranteed physically plausible models are preferable compared to deep networks for this task. The white-box DiffNEA model solves BiC with only four minutes of data via offline MBRL. All network models fail on this task. 
For MBRL the inherent underfitting of white-box models for real-world systems might only be of secondary concern compared to the detrimental effect of divergence to physically unfeasible states. In addition, we extend the existing methods for identification of physics parameters to systems with maximal coordinates and nonholonomic inequality constraints. The real-world experiments show that this approach is also applicable for real-world systems that include unmodeled physical phenomena, such as cable drives and stiction. In future work, we want to look at grey-box models as well as robust policy optimization.  

\medskip \noindent
\textbf{Grey-box Models} This model representation combines black-box and white-box models to achieve high-fidelity approximations of complex physical phenomena with guaranteed avoidance of impossible state regions.
Currently, various initial variants \cite{nguyen2010using, diffNEA, hwangbo2019learning, allevato2020tunenet} exist, but a principled method that optimizes the black- and white-box parameters simultaneously remains an open question.  

\medskip \noindent
\textbf{Robust Policy Optimization} To improve the transferability of the learned optimal policies, robustness w.r.t. to model uncertainty needs to be incorporated into the policy optimization. Within this work we did not incorporate robustness in the policy optimization but plan to extend the DiffNEA model to probabilistic DiffNEA models with domain randomization \cite{ramos2019bayessim, muratore2019assessing, chebotar2019closing}, which is only applicable to white-box models.

\bibliography{references}  
\bibliographystyle{IEEEtran}

\appendix

\begin{figure*}[t]
    \centering
    \includegraphics[width=\textwidth]{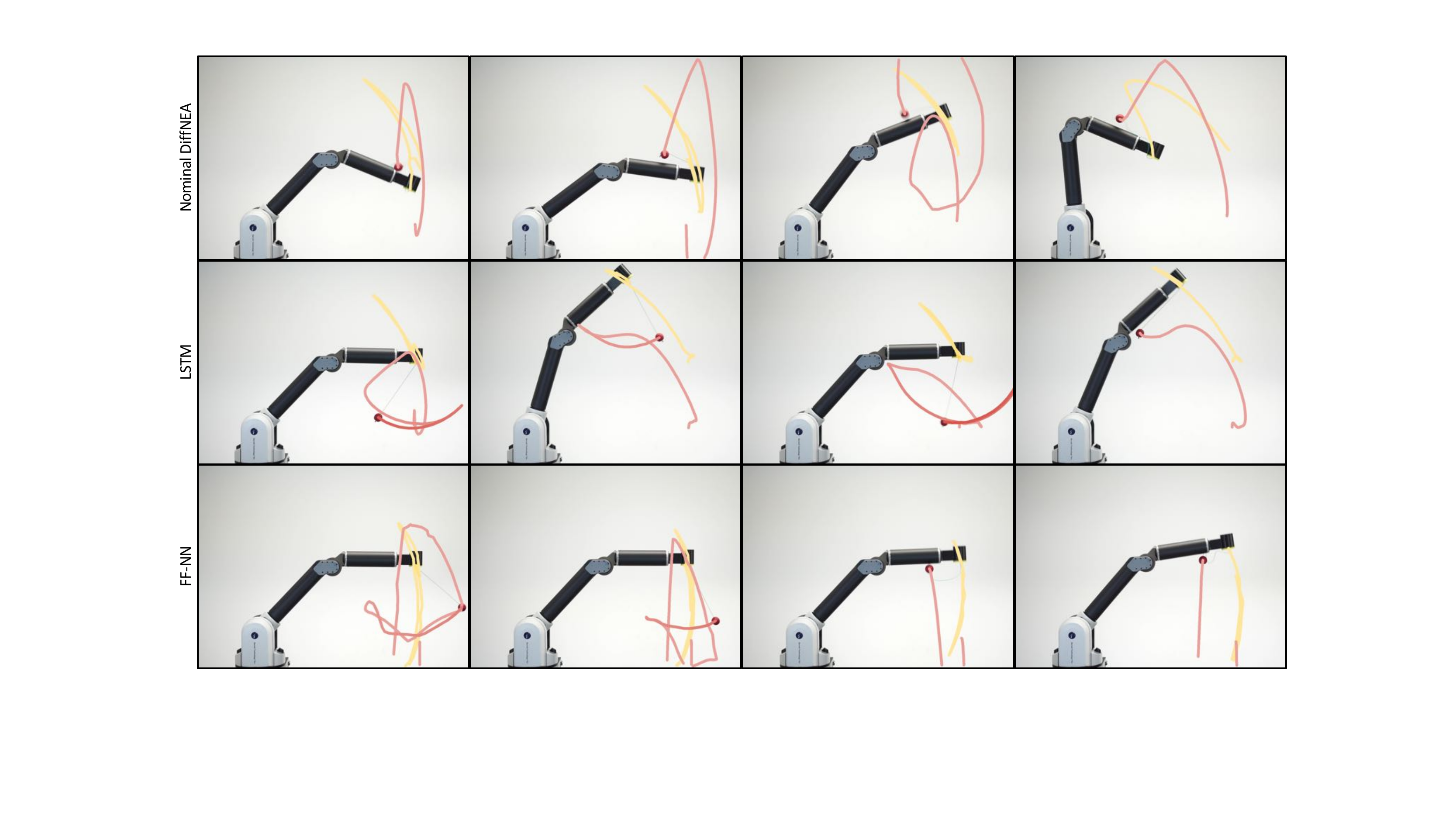}
    \caption{Four different swing-ups for the $40$cm string using the nominal DiffNEA white-box model as well as the LSTM and feed forward neural network (FF-NN) black-box model. Nominal DiffNEA model refers to the DiffNEA model initialized with the measured physical parameters. The swing-ups are learned with eREPS via offline model-based reinforcement learning. None of the different models achieve a successful swing-up despite being evaluated on many reinforcement learning seeds.}
    \vspace{-1.75em}
    \label{fig:baselines}
\end{figure*}

\section*{Newton Euler Equations in Lie Algebra}
\noindent The equation of motions of rigid body trees can be expressed using the Newton Euler equation and Lie Algebra. This formulation is much more compact compared to the traditional formulation of classical textbooks \cite{Featherstone2007rigid}. The pseudocode for the Articulated Body Algorithm (ABA), computing the forward dynamics, can be found in Algorithm \ref{alg:articulated-rigid-body}. The pseudocode for the Recursive Newton Euler Algorithm (RNEA), computing the inverse dynamics, can be found in Algorithm \ref{alg:rnea}. Both algorithms are adapted from \cite{kim2012lie}.
\begin{algorithm}
 \KwIn{
 $\vq_{1:n}$, $\dot{\vq}_{1:n}$, $\vu_{1:n}$ 
}
\KwResult{$\ddot{q}_{0:n}$, $\bar{\va}_{0:n}$, $\bar{\vf}_{0:n}$}
\DontPrintSemicolon
\For{$i\gets1$ \KwTo $n$}
{
// Forward Kinematics\;
$\bar{\vv}_i = Ad_{T^{-1}_{\lambda, i}}\bar{\vv}_{\lambda} {+} \vs_i \dot{q}_i$ \;
$\bar{\bm{\eta}}_i = ad_{\bar{\vv}_i} \vs_i \: \dot{q}_i$
}

\For{$i\gets n$ \KwTo $1$}
{
// Compute lumped inertia \;
$\bar{\mM}_{i:n} {=} \bar{\mM}_i {+} \sum_{k \in \mu} Ad_{T_{i,k}^{-1}}^* \bar{\Pi}_k Ad_{T_{i,k}^{-1}}$\;
// Compute bias forces \;
$\bar{\vf}_{b, i} = - ad_{\bar{\vv}_i}^* \bar{\mM}_{i:n} \bar{\vv}_i {+} \sum_{k \in \mu} Ad_{T_{i,k}^{-1}}^* \left( \bar{\vf}_{b, k} {+} \bar{\bm{\beta}}_k \right)$\;
$\bar{\Psi}_i = \left(\vs_i^T \: \bar{\mM}_{i:n} \: \vs_i\right)^{-1}$\;
$\bar{\bm{\Pi}}_i = \bar{\mM}_{i:n} {-} \bar{\mM}_{i:n} \vs_i \bar{\Psi}_i \vs_i^T \bar{\mM}_{i:n}$\;
$\bar{\bm{\beta}}_i = \bar{\mM}_{i:n}
\left(\bar{\bm{\eta}}_i{+}\vs_i \bar{\Psi}_i \left( u_i {-} \vs_i^T \left( \bar{\mM}_{i:n} \bar{\bm{\eta}}_i {+} \bar{\vf}_{b, i} \right) \right) \right)$\;
}
\For{$i\gets1$ \KwTo $n$}
{
// Newton Euler Equations \;
$\ddot{q}_i = \bar{\Psi}_i \left( u_i {-} \vs_i^T \left(\bar{\mM}_{i:n} \left( Ad_{T_{\lambda, i}^{-1}} \bar{\va}_{\lambda} {+} \bar{\eta}_i \right) {-} \bar{\vf}_{b, i} \right)\right)$\;
$\bar{\va}_i = Ad_{T_{\lambda, i}^{-1}} \bar{\va}_{\lambda} {+} \vs_i \ddot{q}_i {+} \bar{\bm{\eta}}_i$\;
$\bar{\vf}_i = \bar{\mM}_{i:n} \bar{\va}_i {+} \bar{\vf}_{b, i}$\;
}
\caption{Forward dynamics with Articulated Rigid Body for a kinematic tree in terms of Lie algebra. $\lambda$ refers to the parent and $\mu$ refers to the child of the $i$th link}
\label{alg:articulated-rigid-body}
\end{algorithm}

\begin{algorithm}
\KwIn{
 $\vq_{1:n}$, $\dot{\vq}_{1:n}$, $\ddot{\vq}_{1:n}$ 
}
\KwResult{ $\vu_{1:n}$, $\bar{\va}_{0:n}$, $\bar{\vf}_{0:n}$}

\DontPrintSemicolon
\For{$i\gets1$ \KwTo $n$}{
// Forward Kinematics\;
$\bar{\vv}_i = Ad_{T^{-1}_{\lambda, i}}\bar{\vv}_{\lambda} {+} \vs_i \dot{q}_i$ \;
$\bar{\va}_i = Ad_{T^{-1}_{\lambda, i}}\bar{\va}_{\lambda} + ad_{\bar{\vv}_i} \vs_i \: \dot{q}_i + \vs_i \: \ddot{q}_i$
}

\For{$i\gets n$ \KwTo $1$}{
// Newton Euler Equations \;
$\bar{\vf}_i = \bar{\mM}_i\bar{\va}_i - ad_{\bar{\vv}_i}^* \bar{\mM}_i \bar{\vv}_i + \sum_{k \in \mu} Ad_{T_{i,k}^{-1}}^* \bar{\vf}_k$ \;
$u_i = \vs_i\tran \bar{\vf}_i$
}
\caption{Inverse dynamics with Recursive Newton Euler for kinematic tree in terms of Lie algebra. $\lambda$ refers to the parent and $\mu$ refers to the child of the $i$th link}
\label{alg:rnea}
\end{algorithm}

\section*{Virtual Physical Parameters}
\noindent To enable learning of the physical parameters via standard gradient descent methods, we carefully parameterize the physical parameters of the algorithm by a set of unrestricted, \emph{virtual} parameters \cite{ting2006bayesian, sutanto2020encoding}. These virtual are necessary as not all link parameters are physically plausible, e.g., the link mass has to be positive. 

\subsection{Kinematic Transformation}\label{sec:kin} \noindent
The transformation $\mT$ maps between two frames, e.g., the joint and the inertial frame. The fixed transform describes the the distance and rotation between both frames and is parametrized by the translation vector $\vp_k$ and the RPY Euler angles $\vtheta_R{=}[\phi_x, \phi_y, \phi_z]^T$. The transformation is then described by 
\begin{align}
    \mT_O &= 
    \begin{bmatrix}
        \mR_z(\phi_z) \mR_y(\phi_y) \mR_x(\phi_x)  & \vp_k \\
        0   & 1
    \end{bmatrix}
\end{align}
where $\mR_a(\phi)$ denotes the rotation matrix corresponding to the rotation by $\phi$ about axis $a$ using the right-hand rule. Note that the rotation matrices about the elementary axis only depend on $\vtheta_R$ through arguments to trigonometric functions. Due to the periodic nature of those functions we obtain a desired unrestricted parameterization. The complete parameters per transform are summarized as $\vtheta_K{=}\{\vtheta_R, \vp_k \}$

\subsection{Inertias} \noindent
For physical correctness, the diagonal rotational inertia $\mJ_p{=}\text{diag}([J_x, J_y, J_z])$ at the body's CoM and around principal axes must be positive definite and need to conform with the triangle inequalities~\cite{traversaro2016identification}, i.e.,
\begin{equation*}
J_x \leq J_y + J_z , \quad 
J_y \leq J_x + J_z , \quad
J_z \leq J_x + J_y \;.
\end{equation*}
To allow an unbounded parameterization of the inertia matrix, we introduce the parameter vector
$\vtheta_L{=}[\theta_{\sqrt{L_1}},\theta_{\sqrt{L_2}},\theta_{\sqrt{L_3}}]\tran$, 
where $L_i$ represents the central second moments of mass of the density describing the mass distribution of the rigid body with respect to a principal axis frame. Then rotational inertia is described by
\begin{align*}
    \mJ_p = \text{diag}(
\theta_{\sqrt{L_2}}^2{+}\theta_{\sqrt{L_3}}^2, \:
\theta_{\sqrt{L_1}}^2{+}\theta_{\sqrt{L_3}}^2, \:
\theta_{\sqrt{L_1}}^2{+}\theta_{\sqrt{L_2}}^2
).
\end{align*}
The rotational inertia is then mapped to the link coordinate frame using the parallel axis theorem described by 
\begin{equation}
    \mJ = \mR_J \mJ_p \mR_J\tran + m[\vp_m][\vp_m]
\end{equation}
with the link mass $m$ and the translation $\vp_m$ from the coordinate from to the CoM. 
The fixed affine transformation uses the same parameterization as described in \ref{sec:kin}. 
The mass of the rigid body is parameterized by $\theta_{\sqrt{m}}$ where $m{=}\theta_{\sqrt{m}}^2$.
Given the dynamics parameters $\vtheta_{I}{=}\{\vtheta_L, \theta_{\sqrt{m}}, \vtheta_J, \vp_m\}$ for each link, the inertia in the desired frame using as well as generalized inertia $\bar{\mM}$ can be computed.

\end{document}